\documentclass[11pt,a4paper]{article}
\usepackage{times,latexsym}
\usepackage{url}
\usepackage[T1]{fontenc}
\usepackage{latexsym}
\usepackage{amsmath}
\usepackage{amssymb}
\usepackage{graphicx}
\usepackage{subcaption}
\usepackage{multirow}
\usepackage{changepage}
\usepackage{hhline}
\usepackage{color}
\usepackage{booktabs}
\usepackage{authblk}

\usepackage[acceptedWithA]{tacl2018v2}
\usepackage{soul}

\usepackage{xspace,mfirstuc,tabulary}

\newif\iftaclinstructions
\taclinstructionsfalse 
\iftaclinstructions
\renewcommand{\confidential}{}
\renewcommand{\anonsubtext}{(No author info supplied here, for consistency with
	TACL-submission anonymization requirements)}
\newcommand{\instr}
\fi

\iftaclpubformat 

\else

\fi

\title{A Computational Framework for Slang Generation}

\author[1]{Zhewei Sun}
\author[1,2]{Richard Zemel}
\author[1,2]{Yang Xu}
\affil[1]{Department of Computer Science, University of Toronto, Toronto, Canada}
\affil[2]{Vector Institute for Artificial Intelligence, Toronto, Canada}
\affil[ ]{\ttfamily \{zheweisun, zemel, yangxu\}@cs.toronto.edu}

\date{}

\begin{document}
	\maketitle
	\begin{abstract}
		Slang is a common type of informal language, but its flexible nature and paucity of data resources present challenges for existing natural language systems. We take an initial step toward machine generation of slang by developing a framework that models the speaker's word choice in slang context. Our framework encodes novel slang meaning by relating the conventional and slang senses of a word while incorporating syntactic and contextual knowledge in slang usage. We construct the framework using a combination of probabilistic inference and neural contrastive learning. We perform rigorous evaluations on three slang dictionaries and show that our approach not only outperforms state-of-the-art language models, but also better predicts the historical emergence of slang word usages from 1960s to 2000s. We interpret the proposed models and find that the contrastively learned semantic space is sensitive to the similarities between slang and conventional senses of words. Our work creates opportunities for the automated generation and interpretation of informal language. 
	\end{abstract}
	
	\section{Introduction}
	
	Slang is a common type of informal language that appears frequently in daily conversations, social media, and mobile platforms. The flexible and ephemeral nature of slang \cite{eble89,landau84} poses a fundamental challenge for computational representation of slang in natural language systems. As of today, slang constitutes only a small portion of text corpora used in the natural language processing (NLP) community, and it is severely under-represented in standard lexical resources \cite{michel11}. Here we propose a novel framework for automated generation of slang with a focus on generative modeling of slang word meaning and choice.
	
	Existing language models trained on large-scale text corpora have shown success in a variety of NLP tasks. However, they are typically biased toward formal language and under-represent slang. Consider the sentence ``I have a feeling he's gonna \_\_\_ himself someday''. Directly applying a state-of-the-art GPT-2~\cite{radford2019} based language infilling model (e.g., \citet{donahue20}) would result in the retrieval of \textit{kill} as the most probable word choice (probability = 7.7\%). However, such a language model is limited and near-insensitive to slang usage, e.g., \textit{ice}---a common slang alternative for {\it kill}---received virtually 0 probability, suggesting that existing models of distributional semantics, even the transformer-type models, do not capture slang effectively, if at all.
	
	\begin{figure}[t!]
		\begin{subfigure}[b]{0.99\linewidth}
			\includegraphics[width=\linewidth]{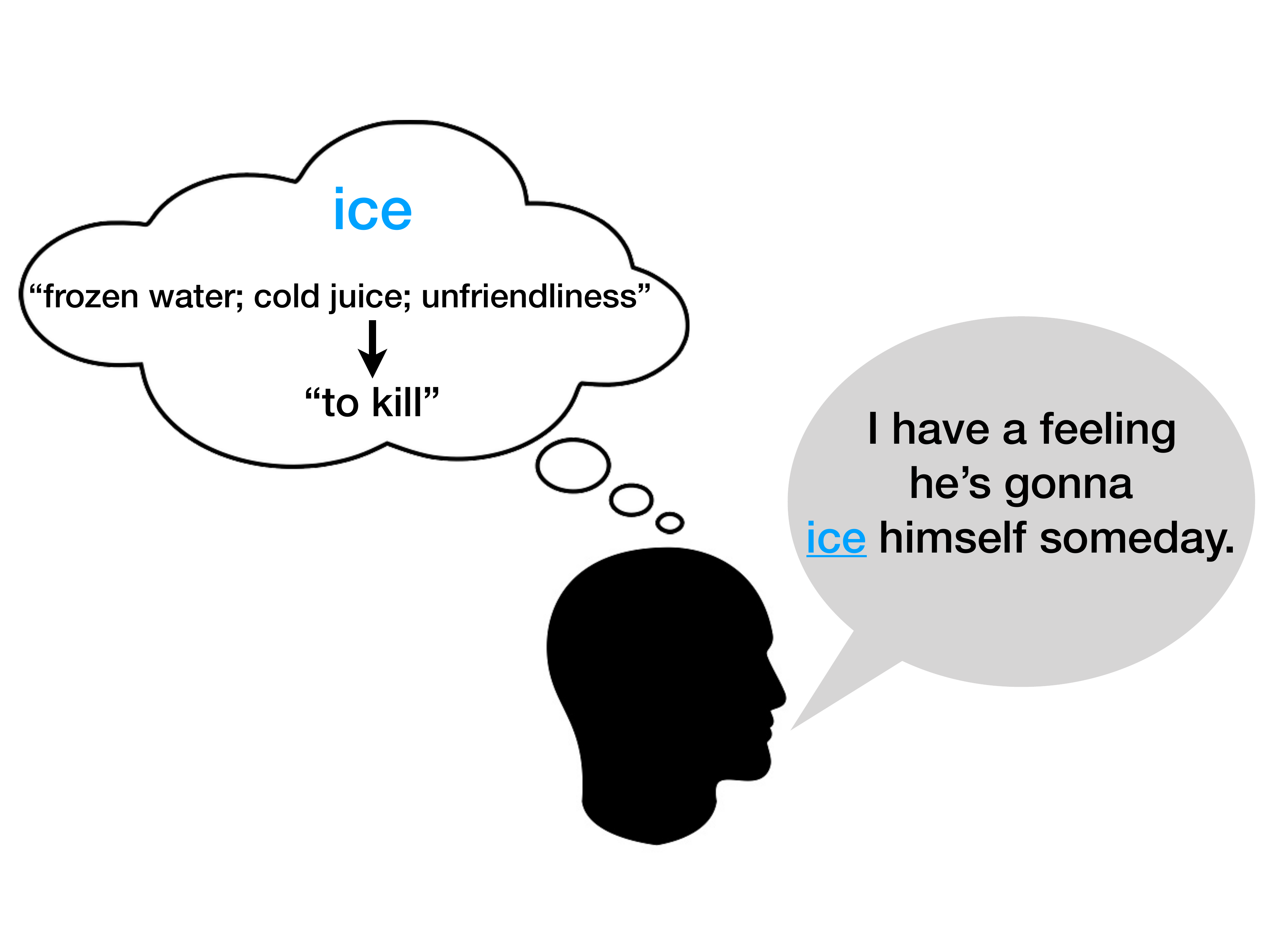}
		\end{subfigure}\vfill
		\caption{A slang generation framework that models speaker's choice of a slang term ({\it ice}) based on the novel sense (``to kill'') in context and relations with conventional senses (e.g., ``frozen water'').} 
		\label{fig1}
	\end{figure}
	
	Our goal is to extend the capacity of NLP systems toward slang in a principled  framework. As an initial step, we focus on modeling the generative process of slang, specifically the problem of slang word choice that we illustrate in Figure~\ref{fig1}. Given an intended slang sense such as ``to kill'', we ask how we can emulate the speaker's choice of slang word(s) in informal context~\footnote{We will use the terms {\it meaning} and {\it sense} interchangeably.}. We are particularly interested in how the speaker chooses existing words from the lexicon and makes innovative use of those words in novel slang context (such as the use of {\it ice} in Figure~\ref{fig1}). 
	
	Our basic premise is that sensible slang word choice depends on linking conventional or established senses of a word (such as ``frozen water'' for {\it ice}) to its emergent slang senses (such as ``to kill'' for {\it ice}). 
	For instance, the extended use of {\it ice} to express killing could have  emerged from the use of cold ice to refrigerate one's remains.
	A principled semantic representation should adapt to such similarity relations. Our proposed framework is aimed at encoding slang that relates informal and conventional word senses, hence capturing semantic similarities beyond those from existing language models. In particular, contextualized embedding models such as BERT would consider ``frozen water'' to be semantically distant or irrelevant from ``to kill'', so they cannot predict {\it ice} to be appropriate for expressing ``to kill'' in slang context.
	
	The capacity for generating novel slang word usages will have several implications and applications. From a scientific view, modeling the generative process of slang word choice will help explain the emergence of novel slang usages over time---we show how our framework can predict the emergence of slang in the history of English. From a practical perspective, automated slang generation paves the way for automated slang interpretation. Existing psycho-linguistic work suggests that language generation and comprehension rely on similar cognitive processes (e.g., \citet{pickering13,renato20}). Similarly, a generative model of slang can be an integral component of slang comprehension that informs the relation between a candidate sense and a query word, where the mapping can be unseen during training. Furthermore, a generative approach to slang may also be applied to downstream tasks such as naturalistic chatbots, sentiment analysis, and sarcasm detection (see work by \citet{shah20} and \citet{wilson20}).
	
	We propose a neural-probabilistic framework that involves three components: 1) a probabilistic choice model that infers an appropriate word for expressing a query slang meaning given its context, 2) an encoder based on contrastive learning that captures slang meaning in a modified embedding space, and 3) a prior that incorporates different forms of context. Specifically, the slang encoder we propose transforms slang and conventional senses of a word into a slang-sensitive embedding space where they will lie in close proximity. As such, senses like ``frozen water'', ``unfriendliness`` and ``to kill'' will be encouraged to be in close proximity in the learned embedding space. Furthermore, the resulting embedding space will also set apart slang senses of a word from senses of other unrelated words, and hence contrasting within-word senses from across-word senses in the lexicon. A practical advantage of this encoding method is that semantic similarities pertinent to slang can be extracted automatically from a small amount of  training data, and the learned semantic space will be sensitive to slang.
	
	Our framework also captures the flexible nature of slang usages in natural context. Here, we focus on syntax and linguistic context, although our framework should allow for the incorporation of social or extra-linguistic features as well. Recent work has found that the flexibility of slang is reflected prominently in syntactic shift \cite{pei19}.   For example, {\it ice}---most commonly used as a noun---is used as a verb to express ``to kill'' (in Figure~\ref{fig1}). We build on these findings by incorporating syntactic shift as a prior in the probabilistic model, which is integrated coherently with the contrastive neural encoder that captures flexibility in slang sense extension. We also show how a contextualized language infilling model can provide additional prior information from  linguistic context (c.f. \citet{erk16}).
	
	To preview our results, we show that our framework yields a substantial improvement on the accuracy of slang generation against state-of-the-art embedding methods including deep contextualized models, in both few-shot and zero-shot settings. We evaluate our framework rigorously on three datasets constructed from slang dictionaries and in a historical prediction task. We show evidence that the learned slang embedding space yields intuitive interpretation of slang and offers future opportunities for informal natural language processing.

	\section{Related Work}
	
	\paragraph{NLP for Non-Literal Language.}
	
	Machine processing of non-literal language has been explored in different linguistic phenomena including metaphor~\cite{shutova13, veale16, gao18, dankers19}, metonymy~\cite{Lapata03, Nissim03, Shutova13_2}, irony~\cite{filatova12}, neologism~\cite{cook10}, idiom~\cite{fazly08, liu18}, vulgarity~\cite{holgate18}, and euphemism~\cite{magu18}.
	Non-literal usages are present in slang, but these existing studies do not directly model the semantics of slang. In addition, work in this area has typically focused on detection and comprehension. In contrast, generation of novel informal language use has been sparsely tackled.
	
	\paragraph{Computational Studies of Slang.}
	
	Slang has been extensively studied as a social phenomenon~\cite{mattiello05}, where social variables such as gender~\cite{blodgett16}, ethnicity~\cite{bamman14}, and social-economic status~\cite{labov72, labov06} have been shown to play important roles in slang construction. More recently, an analysis of social media text has shown that linguistic features also correlate with the survival of slang terms, where linguistically appropriate terms have a higher likelihood of being popularized~\cite{stewart18}.
	
	Recent work in the NLP community has also analyzed slang. \citet{ni17} studied slang comprehension as a translation task.
	In their study, both the spelling of a word and its context are provided as input to a translation model to decode a definition sentence.
	\citet{pei19} proposed end-to-end neural models to detect and identify slang automatically in natural sentences. 
	
	\citet{kulkarni18} have proposed computational models that derive novel word forms of slang from spellings of existing words. 
	Here, we instead explore the generation of novel slang usage from existing words and focus on word sense extension toward slang context, based on the premise that new slang senses are often derived from existing conventional word senses. 
	
	Our work is inspired by previous research suggesting that slang relies on reusing words in the existing lexicon~\cite{eble12}. 
	Previous work has applied cognitive models of categorization to predict novel slang usage~\cite{sun19}. In that work, the generative model is motivated by research on word sense extension~\cite{ramiro18}. In particular, slang generation is operationalized by categorizing slang senses based on their similarities to dictionary definitions of candidate words, 
	enhanced by collaborative filtering~\cite{goldberg92} to capture the fact that words with similar senses are likely to extend to similar novel senses~\cite{lehrer}.  However, this approach presupposes that slang senses are similar to conventional senses of a word represented in standard embedding space, an assumption that is not warranted and yet to be addressed.
	
	Our work goes beyond the existing work in three important aspects: 1) We capture semantic flexibility of slang usage by contributing a novel method based on contrastive learning. Our method encodes slang meaning and conventional meaning of a word under a common embedding space, thereby improving the inadequate existing methodology for slang generation that uses common, slang-insensitive embeddings; 2) We capture syntactic and contextual flexibility of slang usage in a coherent probabilistic framework, an aspect that was ignored in previous work; 3) We rigorously test our framework against slang sense definition entries from three large slang dictionaries and contribute a new dataset for slang research to the community.
	
	\paragraph{Contrastive Learning.}
	
	Contrastive learning is a semi-supervised learning technique used to extract semantic representations in data-scarce situations.
	It can be incorporated into neural networks in the form of twin networks, where two exact copies of an encoder network are applied to two different examples. The encoded representations are then compared and back-propagated. Alternative loss schemes such as Triplet~\cite{Weinberger09, Wang14} and Quadruplet loss~\cite{law13} have also been proposed to enhance stability in training. In NLP,
	contrastive learning has been applied to learn similarities between text~\cite{Mueller16, neculoiu16} and speech utterances~\cite{kamper16} with recurrent neural networks.
	
	The contrastive learning method we develop has two main differences: 1) We do not use recurrent encoders because they perform poorly on dictionary-based definitions; 2) We propose a joint neural-probabilistic framework on the learned embedding space instead of resorting to methods such as nearest-neighbor search for generation.
	
	\section{Computational Framework}
	
	Our computational framework for slang generation comprises three interrelated components: 1) A probabilistic formulation of word choice, extending that in~\citet{sun19} to leverage encapsulated slang senses from a modified embedding space; 2) A contrastive encoder---inspired by variants of twin network~\cite{Baldi93, Bromley94}---that constructs a modified embedding space for slang by adapting the conventional embeddings to incorporate new senses of slang words; 3) A contextually informed prior for capturing flexible uses of naturalistic slang.
	
	\subsection{Probabilistic Slang Choice Model} \label{bayes}
	
	Given a query slang sense $M_S$ and its context $C_S$, we cast the problem of slang generation as inference over candidate words $w$ in our vocabulary. 
	Assuming all candidate words $w$ are drawn from a fixed vocabulary $V$, the posterior is as follows:
	\vspace{-0.15cm}
	\begin{align}
	P(w|M_S, C_S) &\propto P(M_S|w, C_S)P(w|C_S) \nonumber \\
	&\propto P(M_S|w)P(w|C_S) \label{bayeseq}
	\end{align}
	Here, we define the prior $P(w|C_S)$ based on regularities of syntax and/or linguistic context in slang usage (described in Section~\ref{prior}). We formulate the likelihood $P(M_S|w)$\footnote{Here, we only consider linguistically motivated context as $C_S$ and assume the semantic shift patterns of slang are universal across all such contexts.}  by specifying the relations between conventional senses of word $w$ (denoted by $\mathcal{M}_w= \{M_{w_1}, M_{w_2}, \cdots, M_{w_m}\}$, i.e., the set of senses drawn from a standard dictionary) and the query $M_S$ (i.e., slang sense that is outside the standard dictionary). Specifically, we model the likelihood by a similarity function that measures the proximity between the slang sense $M_S$ and the set of conventional senses $\mathcal{M}_w$ of word $w$ in a continuous, embedded semantic space:
	\vspace{-0.6cm}
	\begin{adjustwidth}{-0.2cm}{}
		\begin{align}
		P(M_S|w) &= P(M_S|\mathcal{M}_w) \nonumber \\
		&\propto f(\{sim(E_S, E_{w_i}) ; E_{w_i} \in \mathcal{E}_w \})
		\end{align}
	\end{adjustwidth}
	Here, $f(\cdot)$ is a similarity function in range $[0, 1]$, while $E_S$ and $\mathcal{E}_w$ represent semantic embeddings of the slang sense $M_S$ and the set of conventional senses $\mathcal{M}_w$.	We derive these embeddings from contrastive learning which we describe in detail in Section~\ref{triplet}, and we compare this proposed method with baseline methods that draw embeddings from existing sentence embedding models.
	
	Our choice of the similarity function $f(\cdot)$ is motivated by prior work on few-shot classification. Specifically, we consider variants of two established methods: One Nearest Neighbor (1NN) matching~\cite{koch15, Vinyals16} and Prototypical learning~\cite{snell17}.
	
	The 1NN model postulates that a candidate word should be chosen according to the similarity between the query slang sense and the closest conventional sense:
	\begin{align}
	f_{1nn}(E_S, \mathcal{E}_w) =  \max_{E_{w_i} \in \mathcal{E}_w} sim(E_S, E_{w_i})
	\end{align}
	In contrast, the prototype model postulates that a candidate word should be chosen if its aggregate (or average) sense is in close proximity of the query slang sense:
	\begin{align}
	f_{prototype}(E_S, \mathcal{E}_w) &= sim(E_S, E_w^{prototype}) \nonumber \\
	E_w^{prototype} &= \frac{1}{|\mathcal{E}_w|} \sum_{E_{w_i} \in \mathcal{E}_w} E_{w_i}
	\end{align}
	In both cases, the similarity between two senses is defined by the exponentiated negative squared Euclidean distance in semantic embedding space:
	\begin{equation}
	sim(E_S, E_w) = \exp(-\frac{||E_S - E_w||_2^2}{h_{s}})
	\end{equation}
	Here, $h_s$ is a learned kernel width parameter.
	
	We also consider an enhanced version of the posterior using collaborative filtering \cite{goldberg92}, where words with similar meaning are predicted to shift to similar novel slang meanings. We operationalize this by summing over the close neighborhood of candidate word $L(w)$:
	
	\vspace{-0.2cm}
	\begin{adjustwidth}{-0.3cm}{}
		\begin{align}
		P(w | M_S, C_S)
		& = \sum_{w' \in L(w)} P(w| w')P(w'|M_S, C_S)
		\raisetag{\normalbaselineskip}
		\end{align}
	\end{adjustwidth}
	Here, $P(w'|M_S, C_S)$ is a fixed term calculated identically as in Equation~(\ref{bayeseq}) and $P(w| w')$ is the weighting of words in the close neighborhood of a candidate word $w$. This weighting probability is set proportional to the exponentiated negative cosine distance between $w$ and its neighbor $w'$ defined in pre-trained word embedding space, and the kernel parameter $h_{cf}$ is also estimated from the training data:
		
	\vspace{-0.5cm}
	\begin{adjustwidth}{-0.3cm}{}
		\begin{equation} \label{collabeq}
			P(w|w') \propto sim(w, w') = \exp(-\frac{d(w, w')}{h_{cf}})
		\end{equation}
	\end{adjustwidth}
	Here, $d(w, w')$ is the cosine distance between two words in a word embedding space.

	\subsection{Contrastive Semantic Encoding (CSE)} \label{triplet}
	
	We develop a contrastive semantic encoder for constructing a new embedding space representing slang and conventional word senses that do not bear surface similarities. For instance, the conventional sense of {\it kick} such as ``propel with foot'' can hardly be related to the slang sense of {\it kick} such as ``a strong flavor''. The contrastive embedding space we construct seeks to redefine or warp similarities, such that the otherwise unrelated senses will be in closer proximity than they are under existing embedding methods. For example, two metaphorically related senses can bear strong similarity in slang usage, even though they may be far apart in a literal sense.
	
	We sample triplets of word senses as input to contrastive learning, following work on twin networks~\cite{Baldi93, Bromley94, Chopra05, koch15}. We use dictionary definitions of conventional and slang senses to obtain the initial sense embeddings (See Section~\ref{sentenc} for details). Each triplet consists of 1) an anchor slang sense $M_S$, 2) a positive conventional sense $M_P$, and 3) a negative conventional sense $M_N$. The positive sense should ideally be encouraged to lie closely to the anchor slang sense (in the resulting embedding space), whereas the negative sense should ideally be further away from both the positive conventional and anchor slang senses. Section~\ref{csample} describes the detailed sampling procedures.

	Our triplet network  uses a single neural encoder $g$ to project each word sense representation into a joint embedding space in $\mathbb{R}^d$. 
	\begin{equation}
	E_S = g(M_S); E_P = g(M_P); E_N = g(M_N)
	\end{equation}
	We choose a 1-layer fully connected network with ReLU~\cite{Nair10} as the encoder $g$ for pre-trained word vectors (e.g. fastText). For contextualized embedding models we consider, $g$ will be a Transformer encoder~\cite{vaswani17} . In both cases, we apply the same encoder network $g$  to each of the three inputs. We train the triplet network using the max-margin triplet loss~\cite{Weinberger09}, where the squared distance between the positive pair is constrained to be closer than that of the negative pair with a margin $m$:
	\vspace{-0.5cm}
	\begin{adjustwidth}{-0.2cm}{}
		\begin{align}
		L_{triplet} &= \Big[ m+||E_S - E_P||^2_2 - ||E_S - E_N||^2_2 \Big]_+
		\raisetag{0.15\normalbaselineskip}
		\end{align}
	\end{adjustwidth}
	
	\vspace{-0.8cm}
	\subsection{Triplet Sampling} \label{csample}
	
	To train the triplet network, we build data triplets from every slang lexical entry in our training set. For each slang sense $M_S$ of word $w$, we create a positive pair with each conventional sense $M_{w_i}$ of the same word $w$. Then for each positive pair, we sample a negative example every training epoch by randomly selecting a conventional sense $M_{w'}$ from a word $w'$ that is sufficiently different from $w$, such that the corresponding definition sentence $D_{w'}$ has less than 20\% overlap in the set of content words compared to $M_S$ and any conventional definition sentence $D_{w_i}$ of word $w$. We rank all candidate words in our vocabulary against $w$ by computing cosine distances from pre-trained word embeddings and consider a word $w'$ to be sufficiently different if it is not in the top 20 percent.
	
	\vspace{0.2cm}
	\noindent
	{\bf Neighborhood Sampling (NS).} In addition to using conventional senses of the matching word $w$ for constructing positive pairs, we also sample positive senses from a small neighborhood $L(w)$ of similar words. This sampling strategy provides linguistic knowledge from parallel semantic change to encourage neighborhood structure in the learned embedding space. Sampling from neighboring words also augments the size of the training data considerably in this data-scarce task. We sample negative senses in a similar way, except that we also consider all conventional definition sentences from neighboring words when checking for overlapping senses.
	
	\subsection{Contextual Prior} \label{prior}

	The final component of our framework is the prior $P(w|C_S)$ (see Equation~(\ref{bayeseq})) that captures flexible use of slang words with regard to syntax and distributional semantics. For example, slang exhibits flexible Part-of-Speech (POS) shift, e.g., noun$\rightarrow$verb transition as in the example {\it ice}, and surprisals in linguistic context, e.g., {\it ice} in ``I have a feeling he’s gonna [blank] himself someday.'' Here, we formulate the context $C_S$ in two forms: 1) a syntactic-shift prior, namely the POS information $P_S$ to capture syntactic regularities in slang, and/or 2) a linguistic context prior, namely the linguistic context $K_S$ to capture distributional semantic context when this is available in the data.
	
	\paragraph{Syntactic-Shift Prior (SSP).}
	
	Given a query POS tag $P_S$, we construct the syntactic prior by comparing POS distribution $\mathcal{P}_w$ from literal natural usage of a candidate word $w$ with a smoothed POS distribution $\mathcal{P}_S$ centered at $P_S$.
	However, we cannot directly compare $\mathcal{P}_S$ to $\mathcal{P}_w$ because slang usage often involves shifting POS~\cite{eble12, pei19}. To account for this, we apply a transformation $T$ by counting the number of POS transitions for each slang-conventional definition pair in the training data. Each column of the transformation matrix $T$ is then normalized, so column $i$ of $T$ can be interpreted as the expected slang-informed POS distribution given the $i$'th POS tag in conventional context (e.g., the noun column gives the expected slang POS distribution of a word that is used exclusively as a noun in conventional usage). The slang-contextualized POS distribution $\mathcal{P}_S^*$ can then be computed by applying $T$ on $\mathcal{P}_S$: $\mathcal{P}_S^* = T \times \mathcal{P}_S$. The prior can be estimated by comparing the POS distributions $\mathcal{P}_w$ and $\mathcal{P}_S^*$ via Kullback-Leibler (KL) divergence:
	\begin{equation}
	P(w|C_S) = P(w|P_S) \propto \exp\Big(-KL(\mathcal{P}_w, \mathcal{P}_S^*)\Big)^{\frac{1}{2}}
	\end{equation}
	Intuitively, this prior captures the regularities of syntactic shift in slang usage, and it favors candidate words with POS characteristics that fits well with the queried POS tag in a slang context.
	
	\paragraph{Linguistic Context Prior (LCP).}
	
	We use a language model $P_{LM}$ to a given linguistic context $K_S$ to estimate the probability of each candidate word:
	\begin{equation}
	P(w|C_S) = P(w|K_S) \propto P_{LM}(w|K_S) + \alpha
	\end{equation}
	Here, $\alpha$ is a Laplace smoothing constant. We use the GPT-2 based language infilling model from \citet{donahue20} as $P_{LM}$ and discuss the implementation in Section~\ref{lmbaseline}.

	\section{Experimental Setup}
	
	\subsection{Lexical Resources} \label{data}
	
	We collected lexical entries of slang and conventional words/phrases from three separate online dictionaries:\footnote{We obtained written permissions from all authors for the datasets that we use for this work.} 1) Online Slang Dictionary (OSD),\footnote{OSD: http://onlineslangdictionary.com} 2) Green's Dictionary of Slang (GDoS)~\cite{green10},\footnote{GDoS: https://greensdictofslang.com} and 3) an open source subset of Urban Dictionary (UD) data from Kaggle.\footnote{UD: https://www.kaggle.com/therohk/urban-dictionary-words-dataset}
	In addition, we gathered dictionary definitions of conventional senses of words from the online version of Oxford Dictionary (OD).\footnote{OD: https://en.oxforddictionaries.com}
	
	\paragraph{Slang Dictionary.} \label{secosd}
	
	Both slang dictionaries (OSD and GDoS) are freely accessible online and contain slang definitions with meta-data such as Part-of-Speech tags. 
	Each data entry contains the word, its slang definition, and its part-of-speech (POS) tag. In particular, OSD includes example sentence(s) for each slang entry which we leverage as linguistic context, and GDoS contains time-tagged references that allow us to perform historical prediction (described later). We removed all acronyms (i.e., fully capitalized words) as they generally do not extend meaning, and slang definitions that share more than 50\% content words with any of their corresponding conventional definitions to account for conventionalized slang. We also removed slang with novel word forms where no conventional sense definitions are available.
	Slang phrases were treated as unigrams because our task only concerns the association between senses and lexical items.
	Each sense definition was considered a data point during both learning and prediction. We later partitioned definition entries from each dataset to be used for training, validation, and testing. Note that a word may appear in both training and testing but the pairing between word senses are unique (See Section~\ref{erroranalysis} for discussion).
	
	\paragraph{Conventional Word Senses.}
	
	We focused on the subset of OD containing word forms that are also available in the slang datasets described. For each word entry, we removed all definitions that have been tagged as \textit{informal} because these are likely to represent slang senses. 
	This results in 10,091 and 29,640 conventional sense definitions corresponding to the OSD and GDoS datasets respectively.

	\paragraph{Data Split.}
	
	We used all definition entries from the slang resources such that the corresponding slang word also exists in the collected OD subset. The resulting datasets (OSD and GDS) had 2,979 and 29,300 definition entries respectively, from 1,635 and 6,540 unique slang words, of which 1,253 are shared across both dictionaries. For each dataset, the slang definition entries were partitioned into a 90\% training set and a 10\% test set. 5\% of the data in the training set were set aside for validation when training the contrastive encoder.
	
	\paragraph{Urban Dictionary.}
	
	In addition to the two datasets described above, 	we provide a third dataset based on Urban Dictionary (UD) that are made available via Kaggle. Unlike the previous two datasets, we are able to make this one publicly available without requiring one to obtain prior permission from the data owners.\footnote{Code and data available at: \url{https://github.com/zhewei-sun/slanggen}} To guard against the crowd-sourced and noisy nature of UD, we ensure quality by keeping definition entries such that 1) it has at least 10 more upvotes than downvotes, 2) the word entry exists in one of OSD or GDoS, and 3) at least one of the corresponding definition sentences in these dictionaries have a 20\% or greater overlap in the set of content words with the UD definition sentence. 
	We also remove entries with more than 50\% overlap in content words with any other UD slang definitions under the same word to remove duplicated senses.
	This results in 2,631 definitions entries from 1,464 unique slang words. The corresponding OD subset has 10,357 conventional sense entries. We find entries from UD to be more stylistically variable and lengthier, with a mean entry length of 9.73 in comparison to 7.54 and 6.48 for OSD and GDoS respectively.

	\subsection{Part-of-Speech Data}
	The natural POS distribution $\mathcal{P}_w$ for each candidate word $w$ is obtained using POS counts from the most recent available decade of the HistWords project~\cite{hamilton16}. For word entries that are not available, mostly phrases, we estimate $\mathcal{P}_w$ by counting POS tags from Oxford Dictionary (OD) entries of $w$.
	
	When estimating the slang POS transformation for the syntactic prior, we mapped all POS tags into one of the following six categories: $\{$verb, other, adv, noun, interj, adj$\}$ for the OSD experiments. For GDS, the tag `interj' was excluded as it is not present in the dataset.
	
	\subsection{Contextualized Language Model Baseline} \label{lmbaseline}

	We considered a state-of-the-art GPT-2 based language infilling model from \citet{donahue20} as both a baseline model and a prior to our framework (on the OSD data where context sentences are available for the slang entries). For each entry, we blanked out the corresponding slang word in the example sentence, effectively treating our task as a cloze task. We applied the infilling model to obtain probability scores for each of the candidate words and apply a Laplace smoothing of 0.001. We fine-tuned the LM infilling model using all example sentences in the OSD training set until convergence. We also experiment with a combined prior where the two priors are combined using element-wise multiplication and re-normalization.

	\subsection{Baseline Embedding Methods} \label{sentenc}

	To compare with and compute the baseline embedding methods $M$ for definition sentences, we used 300-dimensional fastText embeddings~\cite{fasttext} pre-trained with subword information on 600 billion tokens from Common Crawl\footnote{http://commoncrawl.org} as well as 768-dimensional Sentence-Bert (SBERT) \cite{reimers19} encoders pretrained on Wikipedia and fine-tuned on NLI datasets \cite{bowman15, williams18}. The fastText embeddings were also used to compute cosine distances $d(w, w')$ in Eq.~\ref{collabeq}. Embeddings for phrases and the fastText-based sentence embeddings were both computed by applying average pooling to normalized word-level embeddings of all content words. In the case of SBERT, we fed in the original definition sentence.
	
	\subsection{Training Procedures}
	
	We trained the triplet networks for a maximum of 20 epochs using Adam ~\cite{kingma15} with a learning rate of $10^{-4}$ for fastText and $2^{-5}$ for SBERT based models. We preserved dimensions of the input sense vectors for the contrastive embeddings learned by the triplet network (that is, 300 for fastText and 768 for SBERT). We used 1,000 fully-connected units in the contrastive encoder's hidden layer for fastText based models. Triplet margins of 0.1 and 1.0 were used with fastText and SBERT embeddings respectively.
	
	We trained the probabilistic classification framework by minimizing negative log likelihood of the posterior $P(w^*|M_S, C_S)$ on the ground-truth words for all definition entries in the training set. We jointly optimized kernel width parameters using L-BFGS-B~\citep{byrd95}. 
	To construct a word $w$'s neighborhood $L(w)$ in both collaborative filtering and triplet sampling, we considered the 5 closest words in cosine distances of their fastText embeddings.

	\section{Results}
	
	\subsection{Model Evaluation}
	
	We first evaluated our models quantitatively by predicting slang word choices: Given a novel slang sense (a definition taken from a slang dictionary) and its part-of-speech, how likely is the model to predict the ground-truth slang recorded in the dictionary? To assess model performance, we allowed each model to make up to $|V|$ ranked predictions where $V$ is the vocabulary of the dataset being evaluated, and we used standard Area-Under-Curve (AUC) percentage from Receiver-Operator Characteristic (ROC) curves to assess overall performance.
	
\begin{figure}[t!]
		\begin{subfigure}[b]{0.99\linewidth}
			\includegraphics[width=\linewidth]{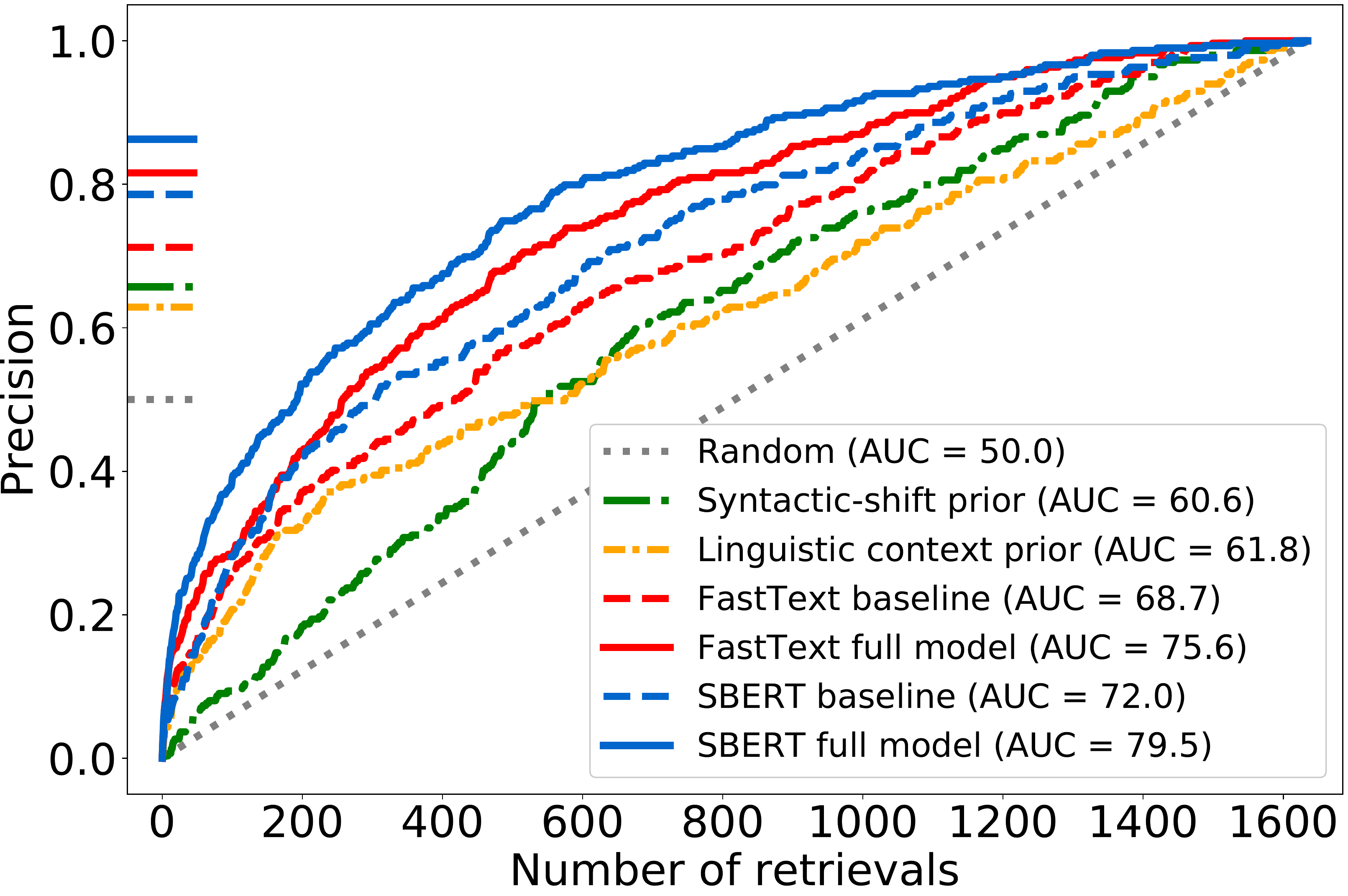}
		\end{subfigure}\vfill
		\caption{ROC curves for slang generation in OSD test set. Collaborative-filtering prototype model was used for prediction. Ticks on the y-axis indicate median precision of the models.} 
		\label{aucfig}
	\end{figure}

	We show the ROC curves for the OSD evaluation in Figure~\ref{aucfig} as an illustration. The AUC metric is similar to and a continuous extension to an F1 score by comprehensively sweeping through the number of candidate words a model is allowed to predict. We find this metric to be the most appropriate because multiple words may be  appropriate to express a probe slang sense.

	To examine the effectiveness of the contrastive embedding method, we varied the semantic representation as input to the models by considering both fastText and SBERT (described in Sec~\ref{sentenc}). For both embeddings, we experimented with the baseline variant without the contrastive encoding (e.g., vanilla embeddings from fastText and SBERT). We then augmented the models incrementally with the contrastive encoder and the priors whenever applicable to examine their respective and joint effects on model performance in slang word choice prediction. We observed that, under both datasets, models leveraging the contrastively learned sense embeddings more reliably predict the ground-truth slang words, indicated by both higher AUC scores and consistent improvement in precision over all retrieval ranks.  Note that the vanilla SBERT model, despite being a much larger model trained on more data, only presented minor performance gains when compared with the plain fastText model. This suggests that simply training larger models on more data does not better encapsulate slang semantics.
	
	We also analyzed whether the contrastive embeddings are robust under different choices of the probabilistic models. Specifically, we considered the following four variants of the models: 1) 1-Nearest Neighbor (1NN),  2) Prototype, 3) 1NN with collaborative filtering (CF), and 4) Prototype with CF. Our results show that applying contrastively learned semantic embeddings consistently improves predictive accuracy across all probabilistic choice models. The complete set of results for all 3 datasets is summarized in Table~\ref{maintable}.
	
	We noted that the syntactic information from the prior improves predictive accuracy in all settings, while by itself predicting significantly better than chance. On OSD, we used the context sentences alone in a contextualized language infilling model for prediction and also incorporating it as a prior. Again, while the prior consistently improves model prediction, both by itself and when paired with the syntactic-shift prior, the language model alone is not sufficient.
	
	We found the syntactic-shift prior and linguistic context prior to be capturing complementary information (mean Spearman correlation of 0.054 $\pm$ 0.003 across all examples), resulting in improved performance when they are combined together. 
	
	However, the majority of the performance gain is attributed to the augmented contrastive embeddings,  which highlights the importance and supports our premise that encoding of slang and conventional senses is crucial to slang word choice.

	\begin{table*}[t!]
		\centering\makebox[\textwidth]{
			\begin{tabular}{lp{0.05cm}rrrr}
				\multicolumn{2}{l}{ Model}&1NN&Prototype&1NN+CF& Proto+CF\\
				\hline
				\addlinespace[0.05cm]
				\multicolumn{6}{l}{Dataset 1: Online Slang Dictionary (OSD)}\\
				\addlinespace[0.05cm]
				\addlinespace[0.1cm]
				Prior Baseline - Uniform &&\multicolumn{4}{c}{51.9}\\
				Prior Baseline - Syntactic-shift &&\multicolumn{4}{c}{60.6}\\
				Prior Baseline - Linguistic Context~\cite{donahue20}   &&\multicolumn{4}{c}{61.8}\\
				Prior Baseline - Syntactic-shift + Linguistic Context  &&\multicolumn{4}{c}{\textbf{67.3}}\\
				\addlinespace[0.15cm]
				FastText Baseline~\cite{sun19} && 63.2&65.2&66.0&68.7\\
				FastText + Contrastive Semantic Encoding (CSE)  &&71.7&71.6&73.0&72.6\\
				FastText + CSE + Syntactic-shift Prior (SSP)  &&73.8&73.4&75.2&74.4\\
				FastText + CSE + Linguistic Context Prior (LCP)  &&73.6&73.2&74.7&73.9\\
				FastText + CSE + SSP + LCP  &&\textbf{75.4}&\textbf{74.9}&\textbf{76.5}&\textbf{75.6}\\
				\addlinespace[0.15cm]
				SBERT Baseline&&67.4&68.1&69.5&72.0\\
				SBERT + CSE  &&76.6&77.4&77.4&78.0\\
				SBERT + CSE + SSP  &&77.6&78.0&78.8&78.9\\
				SBERT + CSE + LCP &&77.8&78.4&78.1&78.7\\
				SBERT + CSE + SSP + LCP  &&\textbf{78.5}&\textbf{79.0}&\textbf{79.4}&\textbf{79.5}\\
				\addlinespace[0.1cm]
				\hline
				\addlinespace[0.05cm]
				\multicolumn{6}{l}{Dataset 2: Green's Dictionary of Slang (GDoS)}\\
				\addlinespace[0.05cm]
				\addlinespace[0.1cm]
				Prior Baseline - Uniform &&\multicolumn{4}{c}{51.5}\\
				Prior Baseline - Syntactic-shift &&\multicolumn{4}{c}{\textbf{61.0}}\\
				\addlinespace[0.15cm]
				FastText Baseline~\cite{sun19} &&68.2&69.9&67.8&69.7\\
				FastText + Contrastive Semantic Encoding (CSE)  &&73.4&74.0&74.1&74.8\\
				FastText + CSE + Syntactic-shift Prior (SSP)  &&\textbf{74.5}&\textbf{74.8}&\textbf{75.2}&\textbf{75.8}\\
				\addlinespace[0.15cm]
				SBERT Baseline&&67.1&68.0&66.8&67.5\\
				SBERT + CSE  &&77.8&78.2&77.4&77.9\\
				SBERT + CSE + SSP  &&\textbf{78.5}&\textbf{78.7}&\textbf{78.3}&\textbf{78.6}\\
				\addlinespace[0.1cm]
				\hline
				\addlinespace[0.05cm]
				\multicolumn{6}{l}{Dataset 3: Urban Dictionary (UD)}\\
				\addlinespace[0.05cm]
				\addlinespace[0.1cm]
				Prior Baseline - Uniform &&\multicolumn{4}{c}{52.3}\\
				\addlinespace[0.15cm]
				FastText Baseline~\cite{sun19} &&65.2&68.8&67.6&70.9\\
				FastText + Contrastive Semantic Encoding (CSE)  &&\textbf{71.0}&\textbf{72.2}&\textbf{71.5}&\textbf{73.7}\\
				\addlinespace[0.15cm]
				SBERT Baseline&&72.4&71.7&74.0&74.4\\
				SBERT + CSE  &&\textbf{76.2}&\textbf{76.6}&\textbf{77.2}&\textbf{78.8}\\
		\end{tabular}}
		\vspace{-0.15cm}
		\caption{\label{maintable} Summary of model AUC scores ($\%$) for slang generation in 3 slang datasets.}
	\end{table*}
	
	\begin{table}
		\begin{subfigure}[b]{\linewidth}
			
			\small
			\begin{tabular}{lrrr}
				\multicolumn{1}{l}{Decade} & \# Test & Baseline & SBERT+CSE+SSP\\
				\hline
				\addlinespace[0.1cm]
				1960s & 2010 & 67.5 & \textbf{77.4} \\
				1970s & 1757 & 66.3 & \textbf{77.9} \\
				1980s & 1655 & 66.3 & \textbf{78.6} \\
				1990s & 1605 & 66.2 & \textbf{75.4} \\
				2000s & 1374 & 65.9 & \textbf{77.0} \\
			\end{tabular}
		\end{subfigure}\hfill
		
		\caption{Summary of model AUC scores in historical prediction of slang emergence (1960s-2000s). The non-contrastive SBERT baseline and the proposed full model (with contrastive embedding, CSE, and syntactic prior, SSP) are compared using collaborative-filtering Prototype. Models were trained and tested incrementally through time (test sizes shown) and trained initially on 20,899 Green's Dictionary  definitions prior to the 1960s.}
		\label{histresults}
	\end{table}

	\subsection{Historical Analysis of Slang Emergence}

	We next performed a temporal analysis to evaluate whether our model explains slang emergence over time. We used the time tags available in the GDoS dataset and predicted historically emerged slang from the past 50 years (1960s--2000s). For a given slang  entry recorded in history, we tagged its emergent decade using the earliest dated reference available in the dictionary. For each future decade $d$, we trained our model using all entries before $d$ and assessed whether our model can predict the choices of slang words for slang senses that emerged in the future decade. We scored the models on slang words that emerged during each subsequent decade, simulating a scenario where future slang usages are incrementally predicted.
	
	Table~\ref{histresults} summarizes the result from the historical analysis for the non-contrastive SBERT baseline and our full model (with contrastive embeddings), based on the GDoS data. AUC scores are similar to the previous findings but slightly lower for both models in this historical setting. Overall, we find the full model to improve the baseline consistently over the course of history examined and achieve similar performance as in the synchronic evaluation. This provides strong evidence that our framework is robust and has explanatory power over the historical emergence of slang.

	\subsection{Model Error Analysis and Interpretation} \label{erroranalysis}
	
	\paragraph{Few-shot vs Zero-shot Prediction.} We analyze our model errors and note that one source of error stems from whether the probe slang word has appeared during training versus not.
	Here, each candidate word is treated as a class and each slang sense of a word seen in the training set is considered a `shot'. 
	In the few-shot case, although the slang sense in question was not observed in prediction, the model has some {\it a priori} knowledge about its target word and how it has been used in slang context (because a word may have multiple slang senses), thus allowing the model to generalize toward novel slang usage of that word.  In the zero-shot case, the model needs to select a novel slang word (i.e., one that never appeared in training) and hence has no direct knowledge about how that word should be extended in a slang context. Such knowledge must be inferred indirectly, and in this case, from the conventional senses of the candidate words. The model can then infer how words with similar conventional senses might extend to slang context.
	
	\begin{table}[t!]
		\begin{subfigure}[b]{\linewidth}
			\centering
			\caption{Online Slang Dictionary (OSD)}
			\small
			\begin{tabular}{lrr}
				\multicolumn{1}{l}{Model} & Few-shot & Zero-shot \\
				\hline
				\addlinespace[0.1cm]
				Prior - Uniform & 55.1 & 47.1 \\
				Prior - Syntactic-shift & 63.4 & \textbf{56.4} \\
				Prior - Linguistic Context & 72.4 & 45.8 \\
				Prior - SSP + LCP & \textbf{74.7} & \textbf{56.4} \\
				\addlinespace[0.1cm]
				FT Baseline & 68.3 & 69.2 \\
				FT + CSE & 74.8 & 69.4 \\
				FT + CSE + SSP & 76.8 & \textbf{70.9} \\
				FT + CSE + LCP & 76.7 & 69.5 \\
				FT + CSE + SSP + LCP& \textbf{78.7} & \textbf{70.9}  \\
				\addlinespace[0.1cm]
				SBERT Baseline & 72.2 & 71.6 \\
				SBERT + CSE & 78.3 & 77.5 \\
				SBERT + CSE + SSP & 79.3 & \textbf{78.3} \\
				SBERT + CSE + LCP & 79.8 & 77.1 \\
				SBERT + CSE + SSP + LCP& \textbf{80.7} & 77.8 \\
			\end{tabular}
		\end{subfigure}\hfill
		
		\vspace{0.2cm}
		
		\begin{subfigure}[b]{\linewidth}
			\centering
			\caption{Green's Dictionary of Slang (GDoS)}
			\small
			\begin{tabular}{lrr}
				\multicolumn{1}{l}{Model} & Few-shot & Zero-shot \\
				\hline
				\addlinespace[0.1cm]
				Prior - Uniform & 51.8 & 48.1 \\
				Prior - Syntactic-shift & \textbf{61.6} & \textbf{54.8} \\
				\addlinespace[0.1cm]
				FT Baseline & 70.6 & \textbf{61.3} \\
				FT + CSE & 76.3 & 59.2 \\
				FT + CSE + SSP & \textbf{77.3} & 60.7 \\
				\addlinespace[0.1cm]
				SBERT Baseline & 68.3 & 59.6 \\
				SBERT + CSE & 79.0 & 66.8 \\
				SBERT + CSE + SSP & \textbf{79.7} & \textbf{67.7} \\
			\end{tabular}
		\end{subfigure}\hfill
		
		\vspace{0.2cm}
		
		\begin{subfigure}[b]{\linewidth}
			\centering
			\caption{Urban Dictionary (UD)}
			\small
			\begin{tabular}{lrr}
				\multicolumn{1}{l}{Model} & Few-shot & Zero-shot \\
				\hline
				\addlinespace[0.1cm]
				Prior - Uniform & 54.2 & 49.1 \\
				\addlinespace[0.1cm]
				FT Baseline & 68.6 & \textbf{75.0} \\
				FT + CSE & \textbf{76.2} & 69.4\\
				\addlinespace[0.1cm]
				SBERT Baseline & 73.0 & \textbf{76.8} \\
				SBERT + CSE & \textbf{80.6} & 75.6 \\
			\end{tabular}
		\end{subfigure}\hfill
		
		\caption{Model AUC scores ($\%$) for Few-shot and Zero-shot test sets (``CSE'' for contrastive embedding, ``SSP'' for syntactic prior, ``LCP'' for contextual prior, and ``FT'' for fastText).}
		\label{zerofewtable}
	\end{table}
	
	Table~\ref{zerofewtable} outlines the AUC scores of the collaboratively filtered prototype models under few-shot and zero-shot settings. For each dataset, we partitioned the corresponding test set by whether the target word appears at least once within another definition entry in the training data. 
	This results in 179, 2,661, and 165 few-shot definitions in OSD, GDoS and UD respectively, along with 120, 269, 96 zero-shot definitions.
	From our results, we observed that it is more challenging for the model to generalize usage patterns to unseen words, with AUC scores often being higher in the few-shot case. Overall, we found the model to have the most issues handling zero-shot cases from GDoS due to the fine-grained senses recorded in this dictionary, where a word has more slang senses on average (in comparison to the OSD and UD data). This issue caused the models to be more biased towards generalizing usage patterns from more commonly observed words. Finally, the SBERT-based models tend to be more robust towards unseen word-forms, potentially benefiting from their contextualized properties.

	\begin{figure}[t!]
		\begin{center}
			\includegraphics[width=0.95\linewidth]{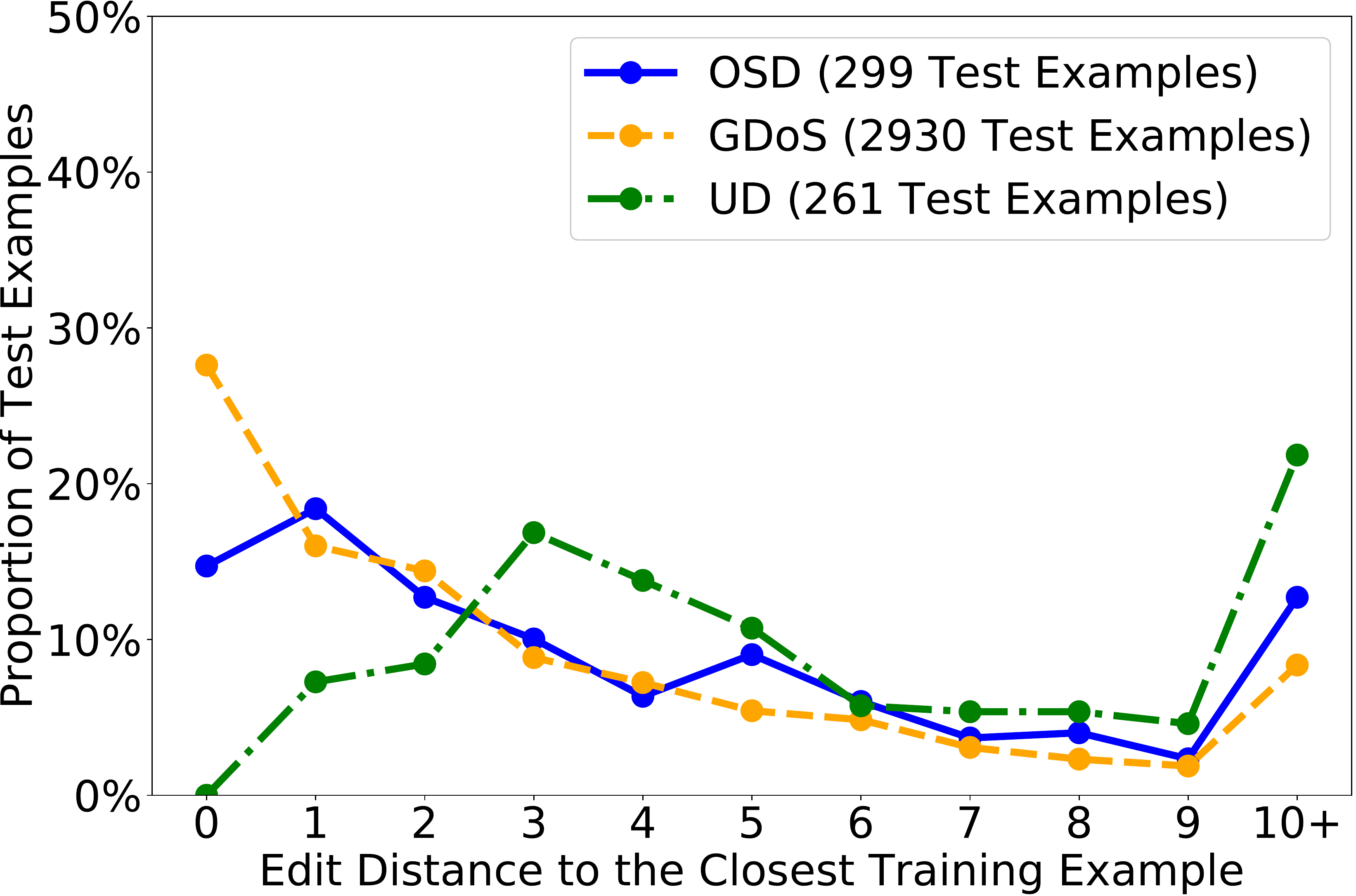}
		\end{center}
		\caption{Degree of synonymy in the test examples relative to training data in each of the 3 datasets.} 
		\label{synmdist}
	\end{figure}

	\begin{figure}[t!]
		\begin{subfigure}[b]{0.95\linewidth}
			\caption{Online Slang Dictionary (OSD)}
			\includegraphics[width=\linewidth]{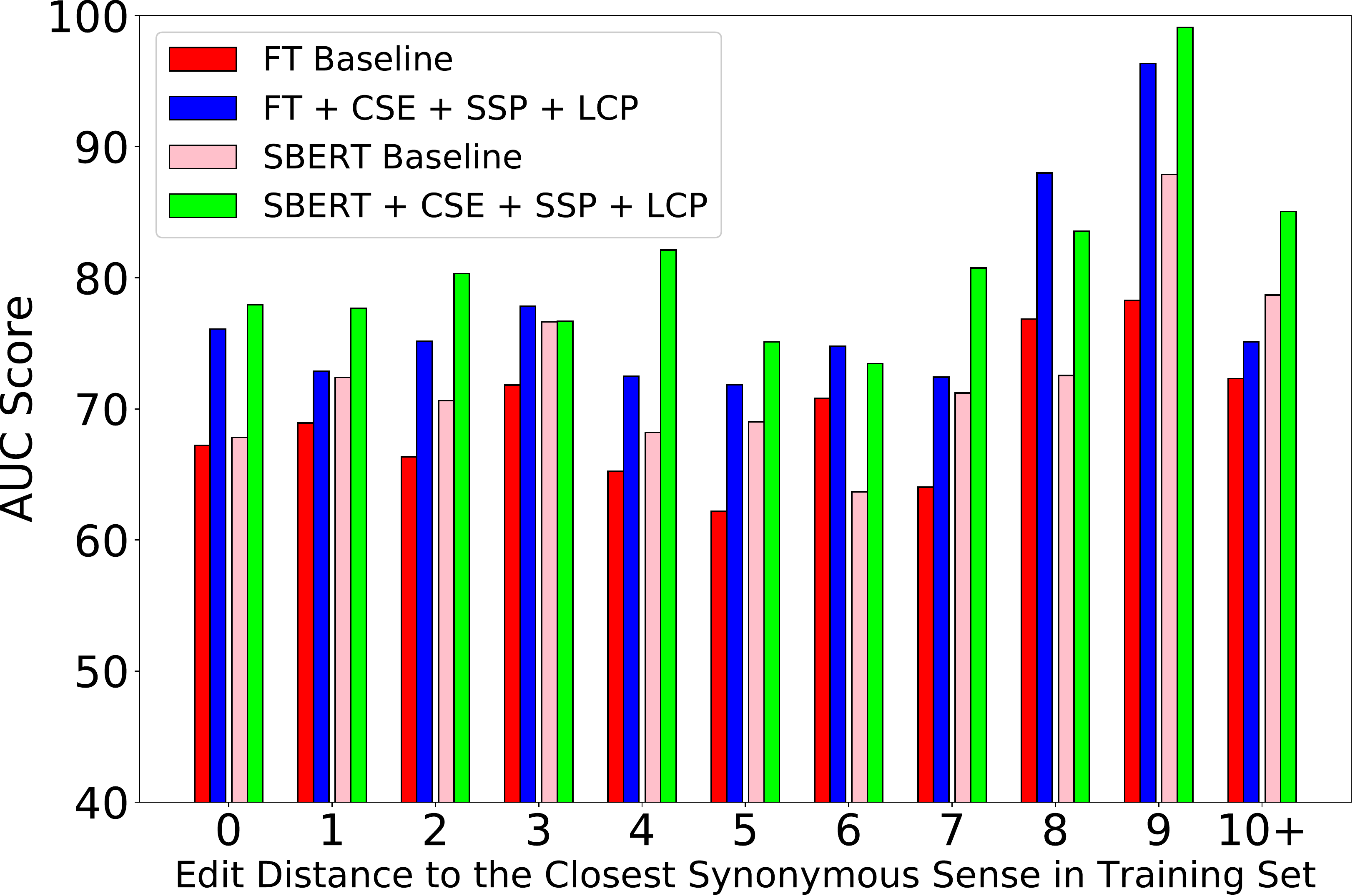}
		\end{subfigure}\
		\begin{subfigure}[b]{0.95\linewidth}
			\vspace{0.2cm}
			\caption{Green's Dictionary of Slang (GDoS)}
			\includegraphics[width=\linewidth]{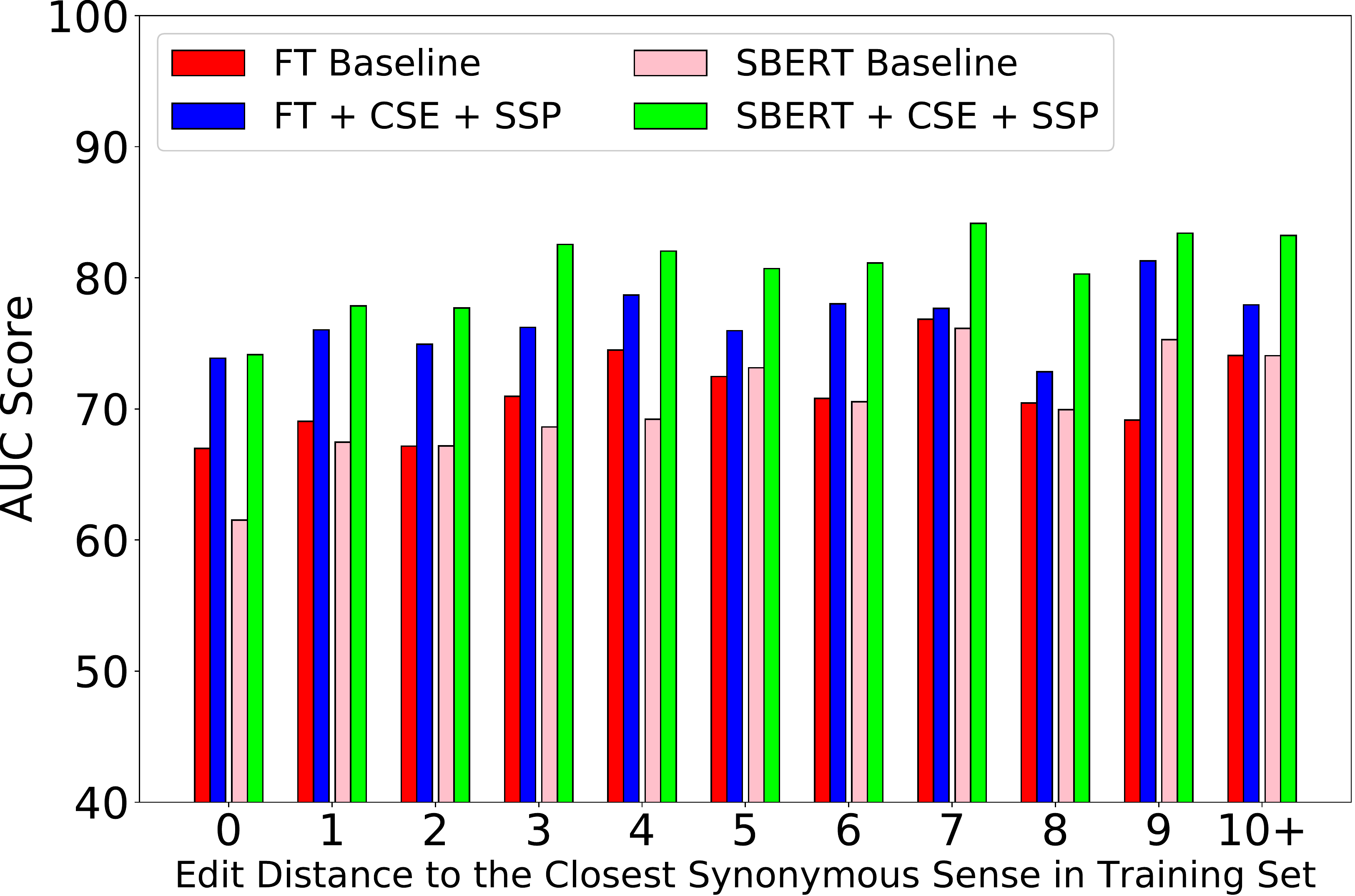}
		\end{subfigure}
		\begin{subfigure}[b]{0.95\linewidth}
			\vspace{0.2cm}
			\caption{Urban Dictionary (UD)}
			\includegraphics[width=\linewidth]{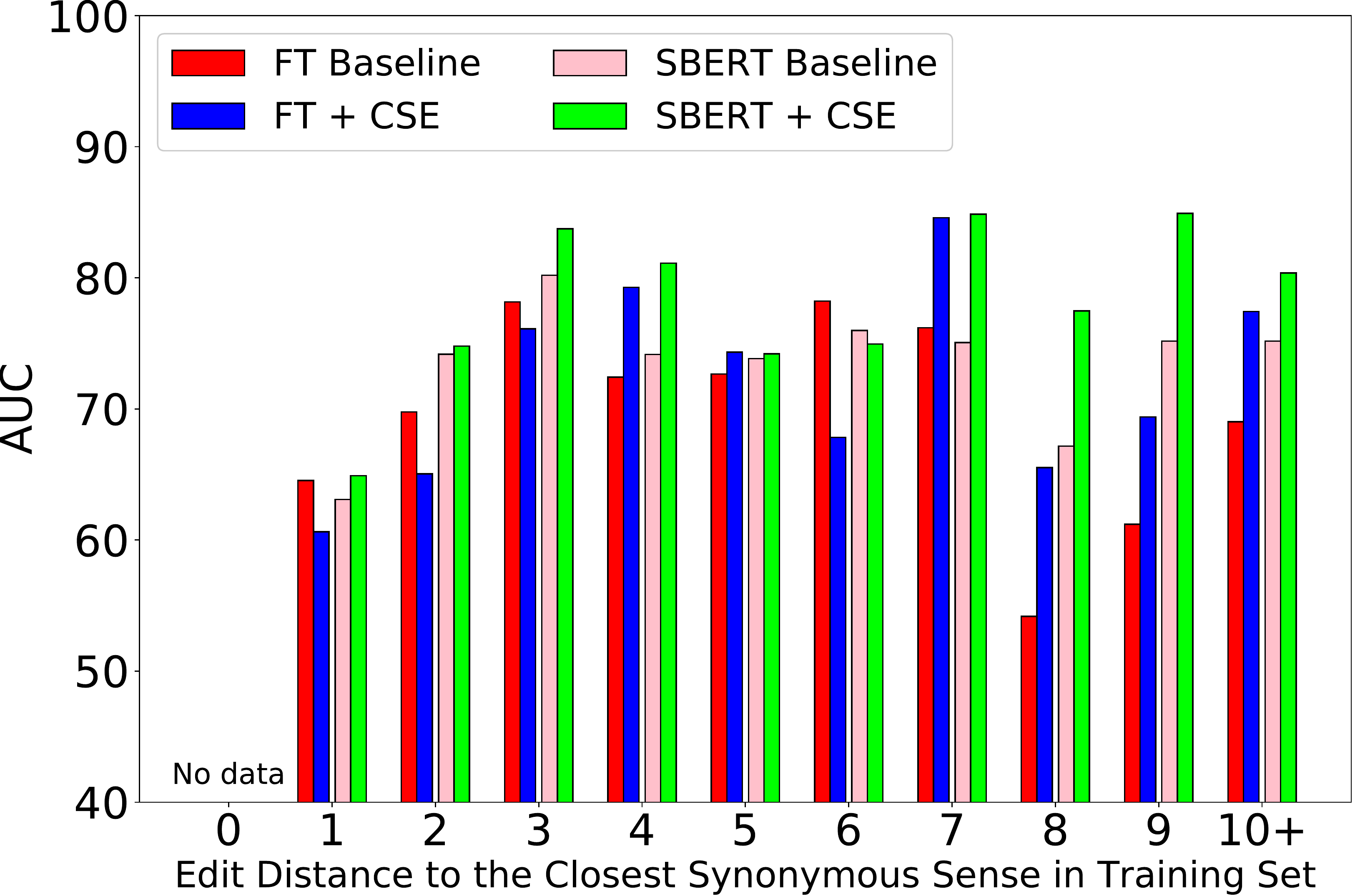}
		\end{subfigure}
		\caption{Model AUC scores ($\%$) under test sets with different degrees of synonymy present in training, for the baselines and the best performing models (under collaborative-filtering prototype).} 
		\label{synmfig}
	\end{figure}

	\begin{table}[t!]
		\begin{subfigure}[b]{\linewidth}
			\centering
			\caption{Online Slang Dictionary (OSD)}
			\small
			\begin{tabular}{lrr}
				\multicolumn{1}{l}{Model} & Training & Testing \\
				\hline
				\addlinespace[0.1cm]
				FT Baseline & 0.33 $\pm$ 0.011 & 0.35 $\pm$ 0.033 \\
				FT + CSE & 0.15 $\pm$ 0.0083 & 0.28 $\pm$ 0.030 \\
				\addlinespace[0.1cm]
				SBERT Baseline & 0.34 $\pm$ 0.011 & 0.32 $\pm$ 0.033 \\
				SBERT + CSE & 0.097 $\pm$ 0.0069 & 0.23 $\pm$ 0.029\\
			\end{tabular}
		\end{subfigure}\hfill
		
		\vspace{0.2cm}
		
		\begin{subfigure}[b]{\linewidth}
			\centering
			\caption{Green's Dictionary of Slang (GDoS)}
			\small
			\begin{tabular}{lrr}
				\multicolumn{1}{l}{Model} & Training & Testing \\
				\hline
				\addlinespace[0.1cm]
				FT Baseline & 0.30 $\pm$ 0.0034 & 0.30 $\pm$ 0.010 \\
				FT + CSE & 0.19 $\pm$ 0.0028 & 0.26  $\pm$ 0.0097 \\
				\addlinespace[0.1cm]
				SBERT Baseline & 0.32  $\pm$ 0.0035 & 0.32  $\pm$ 0.010 \\
				SBERT + CSE & 0.10 $\pm$ 0.0019 & 0.22 $\pm$ 0.0089 \\
			\end{tabular}
		\end{subfigure}\hfill
		
		\vspace{0.2cm}
		
		\begin{subfigure}[b]{\linewidth}
			\centering
			\caption{Urban Dictionary (UD)}
			\small
			\begin{tabular}{lrr}
				\multicolumn{1}{l}{Model} & Training & Testing \\
				\hline
				\addlinespace[0.1cm]
				FT Baseline & 0.34 $\pm$ 0.012 & 0.31 $\pm$ 0.037 \\
				FT + CSE & 0.20 $\pm$ 0.010 & 0.28 $\pm$ 0.033 \\
				\addlinespace[0.1cm]
				SBERT Baseline & 0.34  $\pm$ 0.012 & 0.28 $\pm$ 0.034 \\
				SBERT + CSE & 0.10  $\pm$ 0.0075 & 0.23 $\pm$ 0.031 \\
			\end{tabular}
		\end{subfigure}\hfill
		\caption{Mean Euclidean distance from slang senses to  prototypical conventional senses.}
		\label{distable}
	\end{table}
	
	\paragraph{Synonymous Slang Senses.}
	
	We also examined the influence of synonymy (or sense overlap) in the slang datasets. We quantified the degree of sense synonymy by checking each test sense against all training senses and computing the edit distance between the corresponding sets of constituent content words of the sense definitions.
	
	Figure~\ref{synmdist} shows the distribution of degree of synonymy across all test examples where the edit distance to the closest training example is considered. We perform our evaluation by binning based on the degree of synonymy and summarize the results in Figure~\ref{synmfig}. We do not observe any substantial changes in performance when controlling for the degree of synonymy, and in fact, the highly synonymous definitions appear to be more difficult (as opposed to easier) for the models. Overall, we find the models to yield consistent improvement across different degrees of synonymy, particularly with the SBERT based full model which offers improvement in all cases.

	\begin{table*}[t!]
		\begin{adjustwidth}{-0.1cm}{}
			\centering
			\small
			\begin{tabular}{|p{2.7cm} |p{7.4cm}| p{4.3cm}|}
				\hline
				Model & Top-5 slang words predicted by model & Predicted rank of the true slang\\
				\hline
				
				
				\multicolumn{3}{|l|}{1. True slang: {\it\textbf{kick}}; Slang sense: \textbf{``a thrill, amusement or excitement''} } \\
				\multicolumn{3}{|l|}{\phantom{1 } Sample usage: I got a huge \textit{kick} when things were close to out of hand. } \\
				
				\hline
				
				SBERT Baseline & \textit{thrill, pleasure, frolic, yahoo, sparkle} & \raisebox{-0.2\totalheight}{\includegraphics[width=4.9cm]{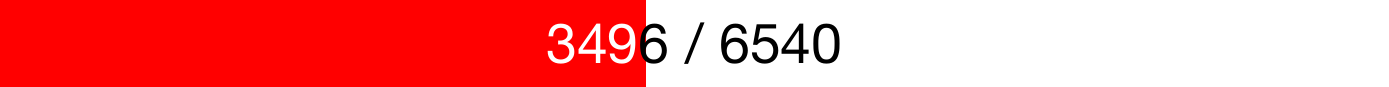}}\\
				\hline
				Full model& \textit{twist, spin, trick, crank, punch} & \raisebox{-0.2\totalheight}{\includegraphics[width=4.9cm]{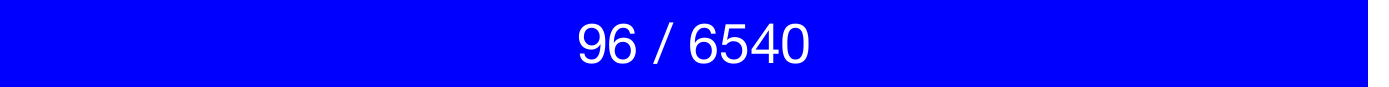}}\\
				\hline
				
				
				\multicolumn{3}{|l|}{2. True slang: {\it\textbf{whiff}}; Slang sense: \textbf{``to kill, to murder, [play on SE, to blow away]''} } \\
				\multicolumn{3}{|l|}{\phantom{1 } Sample usage: The trouble is he wasn’t alone when you \textit{whiffed} him. } \\
				
				\hline
				
				SBERT Baseline & \textit{suicide, homicide, murder, killing, rape} & \raisebox{-0.2\totalheight}{\includegraphics[width=4.9cm]{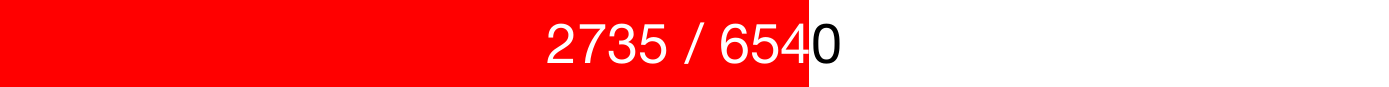}}\\
				\hline
				Full model& \textit{spill, swallow, blow, flare, dash} & \raisebox{-0.2\totalheight}{\includegraphics[width=4.9cm]{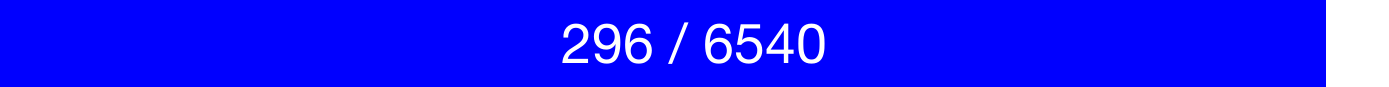}}\\
				\hline
				
				
				\multicolumn{3}{|l|}{3. True slang: {\it\textbf{chirp}}; Slang sense: \textbf{``an act of informing, a betrayal''} } \\
				\multicolumn{3}{|l|}{\phantom{1 } Sample usage: Once we’re sure there’s no back-fire anywhere, the Sparrow will \textit{chirp} his last chirp. } \\
				
				\hline
				
				SBERT Baseline & \textit{dupe, sin, scam, humbug, hocus} & \raisebox{-0.2\totalheight}{\includegraphics[width=4.9cm]{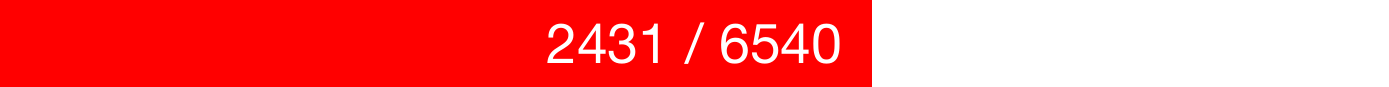}}\\
				\hline
				Full model& \textit{chirp, squeal, squawk, fib, chat} & \raisebox{-0.2\totalheight}{\includegraphics[width=4.9cm]{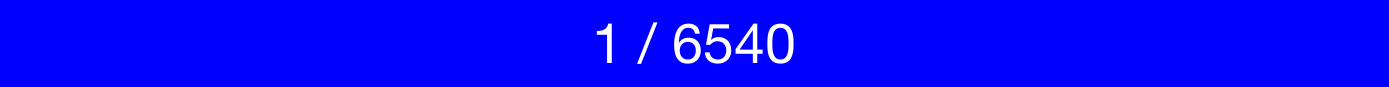}}\\
				\hline

				
				\multicolumn{3}{|l|}{4. True slang: {\it\textbf{red}}; Slang sense: \textbf{``a communist, a socialist or anyone considered to have left-wing leanings''} } \\
				\multicolumn{3}{|l|}{\phantom{1 } Sample usage: Why the hell would I bed a \textit{red}? } \\
				
				\hline
				
				SBERT Baseline & \textit{leveller, wildcat, mole, pawn, domino} & \raisebox{-0.2\totalheight}{\includegraphics[width=4.9cm]{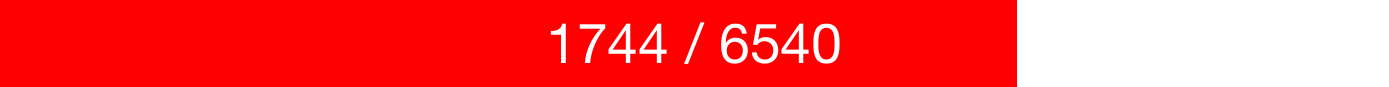}}\\
				\hline
				Full model& \textit{orange, bluey, black and tan, violet, shadow} & \raisebox{-0.2\totalheight}{\includegraphics[width=4.9cm]{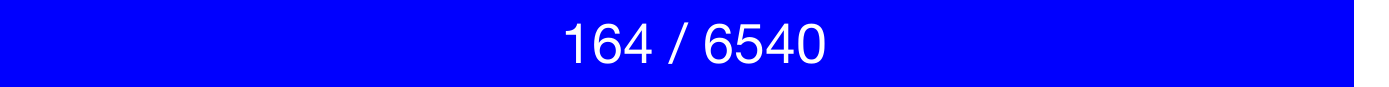}}\\
				\hline
				
				
				\multicolumn{3}{|l|}{5. True slang: {\it\textbf{team}}; Slang sense: \textbf{``a gang of criminals''} } \\
				\multicolumn{3}{|l|}{\phantom{1 } Sample usage: And a little \textit{team} to follow me – all wanted up the yard.} \\
				
				\hline
				
				SBERT Baseline & \textit{gangster, hoodlum, thug, mob, gangsta} & \raisebox{-0.2\totalheight}{\includegraphics[width=4.9cm]{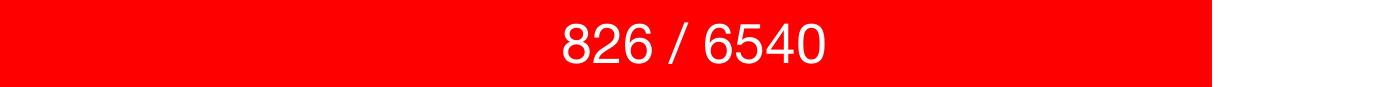}}\\
				\hline
				Full model& \textit{brigade, mob, business, gang, school} & \raisebox{-0.2\totalheight}{\includegraphics[width=4.9cm]{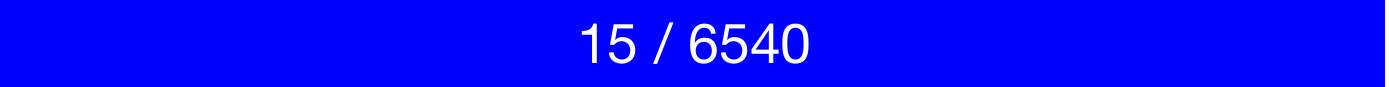}}\\
				\hline
				
			\end{tabular}
			\caption{Example slang word predictions from the contrastively learned full model and SBERT baseline (with no contrastive embedding) on slang usage  from the Green's Dictionary. Each example shows the true slang, the probe slang sense, a sample usage,  the alternative slang words predicted by each model, and the predicted rank (colored bars indicate inverse rank) of the true slang from a lexicon of 6,540 words.
			}
			\label{tableex}
		\end{adjustwidth}
	\end{table*}

	\paragraph{Semantic Distance.}
	
	To understand the consequence of contrastive embedding, we examine the relative distance between conventional and slang senses of a word in embedding space and the extent to which the learned semantic relations might generalize. We measured the Euclidean distance between each slang embedding with the prototype sense vector of all candidate words, without applying the probabilistic choice models. 
	Table~\ref{distable} shows the ranks of the corresponding candidate words, averaged over all slang sense embeddings considered and normalized between 0 and 1. We observed that contrastive learning indeed brings closer slang and conventional senses (from the same word), as indicated by lower mean semantic distance after the embedding procedure is applied. Under both fastText and SBERT, we obtained significant improvement on both the OSD and GDoS test sets ($p < 0.001$). 
	On UD, the improvement is  significant for SBERT ($p = 0.018$) but marginal for fastText ($p=0.087$). 
	
	\paragraph{Examples of Model Prediction.}
	
	Table~\ref{tableex} shows 5 example slangs from the GDoS test set and the top words predicted by both the baseline SBERT model and the full SBERT-based model with contrastive learning. 
	The full model exhibits a greater tendency to choose words that appear  remotely related to the queried sense (e.g., \textit{spill}, \textit{swallow} for the act of killing), while the baseline model favors words that share only surface semantic similarity (e.g., retrieving \textit{murder} and \textit{homicide} directly). We found cases where the model extends meaning metaphorically (e.g., animal to action, in the case of \textit{chirp}), euphemistically (e.g., \textit{spill} and \textit{swallow} for kill), and generalization of a concept (e.g., \textit{brigade} and \textit{mob} for gang), all of which are commonly attested in slang usage \cite{eble12}. 
	
	We found the full model to achieve better retrieval accuracy in cases where the queried slang undergoes a non-literal sense extension, whereas the baseline model is situated at retrieving candidate words with incremental or literal changes in meaning. We also noted many cases where the true slang word is difficult to predict without appropriate background knowledge. For instance, the full-model suggested words such as \textit{orange} and \textit{bluey} to mean ``a communist'' but could not pinpoint the color \textit{red} without knowing its cultural association to communism.  Finally, we observed that our model to perform generally worse when the target slang sense can hardly be related to conventional senses of the target word, suggesting that cultural knowledge may be important to consider in the future.

	\section{Conclusion}
	
	We have presented a framework that combines probabilistic inference with neural contrastive learning to generate novel slang word usages. Our results suggest that capturing semantic and contextual flexibility simultaneously helps to improve the automated generation of slang word choices with limited training data. To our knowledge this work constitutes the first formal computational approach to modeling slang generation, and we have shown the promise of the learned semantic space for representing slang senses. Our framework will provide opportunities for future research in the natural language processing of informal language, particularly the automated interpretation of slang.
	
	\section*{Acknowledgements}
	
	We thank the anonymous TACL reviewers and action editors for their constructive and detailed comments. We thank Walter Rader and Jonathon Green respectively for their permissions to use The Online Slang Dictionary and Green’s Dictionary of Slang for our research. We thank Graeme Hirst, Ella Rabinovich, and members of the Language, Cognition, and Computation (LCC) Group at the University of Toronto for offering thoughtful feedback to this work. We also thank Dan Jurafsky and Derek Denis for stimulating discussion. This work was supported by a NSERC Discovery Grant RGPIN-2018-05872 and a Connaught New Researcher Award to YX.
	
	\bibliography{main-2701-Sun-arxiv}
	\bibliographystyle{acl_natbib}
	
\end{document}